%% file: main.tex
\documentclass[letterpaper, 10 pt, journal, twoside]{IEEEtran}

\input{includes}

\begin{document}

\title{\mnet: A Tool to Generate and Benchmark Motion Planning Datasets}

\author{Constantinos Chamzas$^{1}$, Carlos Quintero-Pe\~na$^{1}$, Zachary Kingston$^{1}$, Andreas Orthey$^{2}$, Daniel Rakita$^{3}$, \\
Michael Gleicher$^{3}$, Marc Toussaint$^{2}$ and Lydia E. Kavraki$^{1}$ 
\thanks{Manuscript received: September, 9, 2021; Accepted November, 18, 2021.
This letter was recommended for publication by Associate Editor Oren Salzman and Editor Hanna Kurniawati upon evaluation of the reviewers’ comments.
At Rice University, this work was supported in part by NSF 1718478, NSF 2008720, NSF-GRFP 1842494, and Rice University Funds.
Work on this project by DR and MG was supported in part by NSF 1830242.} 
\thanks{$^{1}$CC, CQP, ZK, and LEK are with the Department of Computer Science, Rice University, Houston, TX, USA  {\tt \footnotesize \{chamzas,carlosq,zak,kavraki\}@rice.edu}}%
\thanks{$^{2}$AO and MT are with the Learning and Intelligent Systems Lab, TU Berlin, Germany 
  {\tt \footnotesize \{orthey\}@campus.tu-berlin.de, \{toussaint\}@tu-berlin.de}}%
\thanks{$^{3}$DR and MG are with the Department of Computer Sciences, University of Wisconsin-Madison, {\tt \footnotesize \{rakita, gleicher\}@cs.wisc.edu}}%
   \thanks{Digital Object Identifier (DOI): see top of this page.}
}

\markboth{IEEE Robotics and Automation Letters. Preprint Version. Accepted November, 2021}
{Chamzas \MakeLowercase{\textit{et al.}}: \mnet: A Tool to Generate and Benchmark Motion Planning Datasets} 

\maketitle

\begin{abstract}
  \input{src/abstract}
\end{abstract}


\IEEEpeerreviewmaketitle

\input{src/introduction}
\input{src/related}
\input{src/modules}
\input{src/usecases}
\input{src/experiments}

\input{src/discussion.tex}


\bibliographystyle{IEEEtran}
\bibliography{bib/references.bib}

\end{document}

%% file: includes.tex
\usepackage[utf8]{inputenc}

\usepackage{times}
\usepackage{xinttools}
\usepackage{tabularx}

\usepackage{multirow}
\usepackage{xspace}
\usepackage{comment}
\usepackage{graphicx}
\usepackage{xcolor}

\usepackage{xcolor}
\definecolor{gk}{RGB}{80, 80, 80}
\definecolor{gg}{HTML}{5f9411}
\definecolor{gb}{HTML}{417598}
\definecolor{gr}{HTML}{d15120}
\definecolor{gy}{HTML}{d2ad00}

\usepackage{amsmath}
\usepackage{amssymb}
\usepackage{amsthm}

\newtheoremstyle{main}
                {1em}                                                
                {1em}                                                
                {\normalfont}                                        
                {0pt}                                                
                {\scshape}                                           
                {\\*}                                                
                {2pt}                                                
                {\thmname{#1}\thmnumber{ #2}: \thmnote{\itshape #3}} 

\theoremstyle{main}

\usepackage[linesnumbered,ruled,noend]{algorithm2e}

\usepackage[
  activate   = {true},
  protrusion = true,
  expansion  = true,
  kerning    = true,
  spacing    = true,
  tracking   = false,
  auto       = true,
  selected   = true,
  factor     = 1000,
  stretch    = 10,
  shrink     = 25,
]{microtype}

\makeatletter
\let\NAT@parse\undefined
\makeatother
\usepackage[pdfa,colorlinks,bookmarksopen,bookmarksnumbered,allcolors=gg]{hyperref}
\usepackage{cite}
\usepackage[inkscapelatex=false]{svg}

\usepackage{listings}

\newcommand{\mydots}{.\hspace{-.1em}.\hspace{-.1em}.}

\newcommand{\ssampler}{\emph{Scene Sampler}\xspace}
\newcommand{\ogen}{\emph{Octomap Generator}\xspace}
\newcommand{\pgen}{\emph{Problem Generator}\xspace}
\newcommand{\ssetup}{\emph{Setup}\xspace}

\newcommand{\mnet}{\textsc{MotionBenchMaker}\xspace}

\newcommand{\cspace}{\mbox{\ensuremath{\mathcal{C}}-space}\xspace}

\newcommand{\se}{\ensuremath{\textsc{se}(\oldstylenums{3})}\xspace}

\newcommand{\rrtc}{\textsc{rrt}Connect\xspace}
\newcommand{\bkpiece}{\textsc{bkpiece}\xspace}
\newcommand{\biest}{\textsc{biest}\xspace}
\newcommand{\ait}{\textsc{ait*}\xspace}
\newcommand{\bit}{\textsc{bit*}\xspace}
\newcommand{\rrts}{\textsc{rrt*}\xspace}

\newcommand{\xaxis}{\textsc{x-axis}\xspace}
\newcommand{\yaxis}{\textsc{y-axis}\xspace}

\newcommand{\dof}{\textsc{dof}\xspace}



%% file: src/abstract.tex
Recently, there has been a wealth of development in motion planning for robotic manipulation---new motion planners are continuously proposed, each with their own unique strengths and weaknesses.
However, evaluating new planners is challenging and researchers often create their own ad-hoc problems for benchmarking, which is time-consuming, prone to bias, and does not directly compare against other state-of-the-art planners.
We present \mnet, an open-source tool to generate benchmarking datasets for realistic robot manipulation problems.
\mnet is designed to be an extensible, easy-to-use tool that allows users to both generate datasets and benchmark them by comparing motion planning algorithms.
Empirically, we show the benefit of using \mnet as a tool to procedurally generate datasets which helps in the fair evaluation of planners.
We also present a suite of 40 prefabricated datasets, with $5$ different commonly used robots in $8$ environments, to serve as a common ground to accelerate motion planning research.

%% file: src/introduction.tex
\section{Introduction}
\IEEEPARstart{M}{otion} planning is a core component of robotic manipulation~\cite{mason2018toward}.
For example, motion planning is essential in pick-and-place tasks~\cite{Leitner2017acrv}, finding geometrically-constrained motions such as opening drawers and doors~\cite{Kingston2018}, and as a tool in task and motion planners to evaluate the feasibility of long-horizon plans~\cite{Garrett2021}.
The multitude of applications of motion planning has given rise to a multitude of motion planners to tackle these specific problems, each employing their own heuristics~\cite{Orthey2021TRO} to address the challenging general problem~\cite{Canny1988}.

Despite the plethora of planning methods proposed over the years, little emphasis has been placed on creating a common ground to evaluate these planners---there are no shared benchmarking datasets tailored to manipulation problems that are commonly found in the literature~\cite{antonelli2015}.
The lack of shared environments for evaluation often forces researchers to create their own, making it challenging for practitioners to understand the advantages or disadvantages of a particular method if not directly compared.
Additionally, crafting bespoke planning problems to evaluate a method is very time consuming, and could lead to incorrect conclusions due to unintentional biases in design.
Finally, with the advent of learning-based planning methods (e.g.,~\cite{Chamzas2020sampling, Ichter2018, Chen2020}), there has been an increased need for readily available open-source datasets that can be used for training and testing.

\begin{figure}
    \centering
    \includegraphics[width=0.95\linewidth]{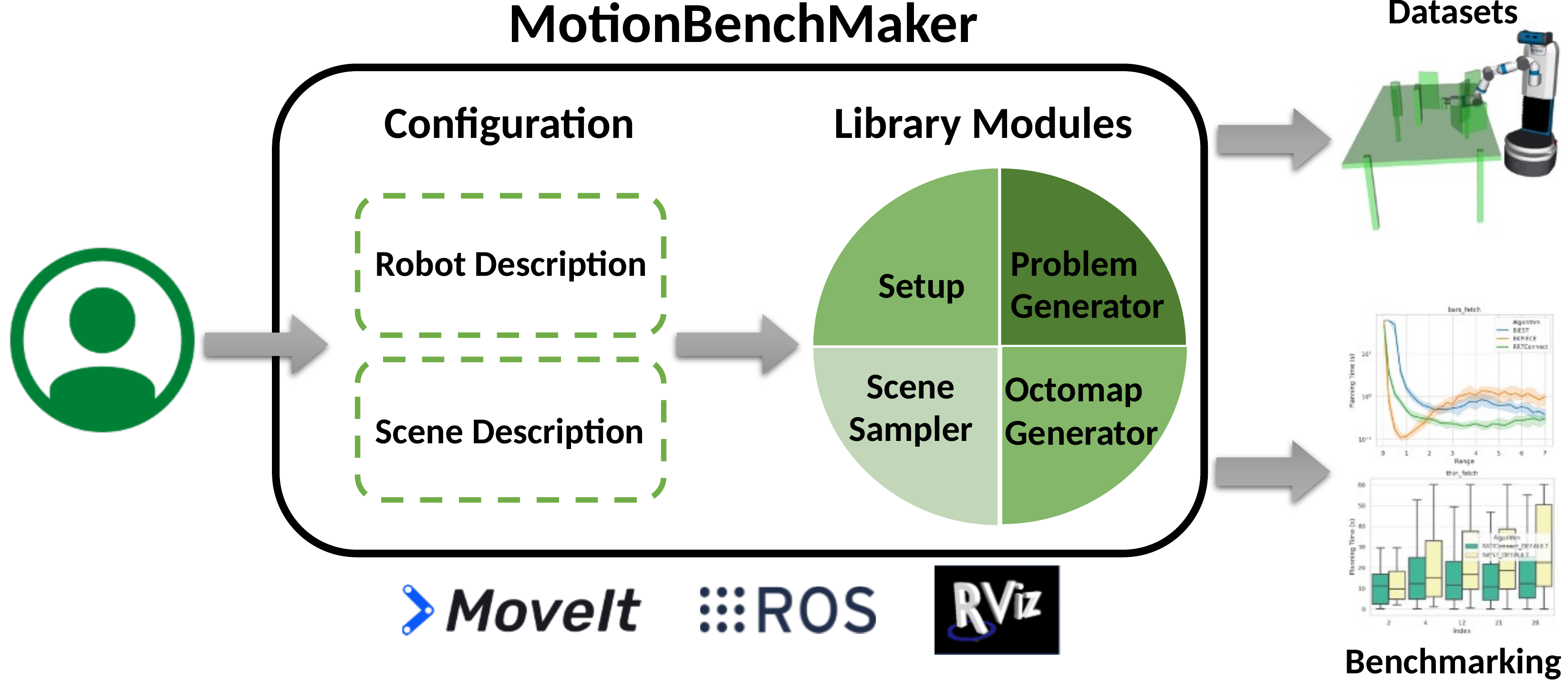}
    \caption{\mnet architecture}
    \label{fig:mnet_arch}
\end{figure}

We introduce \mnet, a tool that facilitates the creation of motion planning datasets to ease the evaluation of motion planning algorithms in ``realistic'' manipulation tasks.
\mnet was inspired by common issues found in evaluating sampling-based planners on high-\dof robots.
Unlike most existing benchmarking resources, which are designed for low-\dof robots or free-flying systems (see~\autoref{tab:related_work}), \mnet is intended for modern high-\dof robots in ``realistic'' scenes, and its capabilities are broadly useful to other types of planners, e.g., classical, optimization-based, and learning-based.
\mnet consists of a set of tools in the form of modules~\autoref{fig:mnet_arch}, which can be utilized by user scripts and human-readable configuration files.
The two main use cases for \mnet are the generation of motion planning datasets and subsequent evaluation of motion planners on these datasets.
We also provide 40 prefabricated datasets ($5$ different robots in $8$ different environments) which are open source along with \mnet\footnote{\url{https://github.com/KavrakiLab/motion_bench_maker}}. A video is also provided that visually presents this work \footnote{\url{https://youtu.be/t96Py0QX0NI}}.

\mnet specifies motion planning problems as \emph{robot-agnostic manipulation queries} which depend only on the environment geometry---with this, it is easy to integrate new problems and new robots to create new datasets (e.g., see~\autoref{fig:problems}).
Planning problems within a dataset are randomly generated given a nominal environment and a set of tunable parameters. These parameters specify how objects in the environment can vary in their pose and control how new samples of planning problems are be procedurally generated.
\mnet also provides the ability to convert scenes described with geometric primitives and meshes to a ``sensed'' representation, i.e., point clouds and octomaps~\cite{Hornung2013}.
\mnet is fully compatible with the ROS~\cite{quigley2009ros} ecosystem of tools and interfaces such as visualization with RViz and motion planning through MoveIt~\cite{MoveIt} and Robowflex~\cite{Robowflex}. 
To summarize, with \mnet we contribute a tool which 
\begin{itemize}
    \item has a modular and open architecture to facilitate creating new datasets,
    \item procedurally generates new datasets by randomly varying scenes,
    \item can convert scenes to ``sensed'' representations,
    \item has benchmarking capabilities,
    \item and is easy to integrate into the existing ROS ecosystem.
\end{itemize}

The rest of the paper is organized as follows.
In~\autoref{sec:related} we review other works in robotic benchmarking and dataset.
In~\autoref{sec:components} we describe the modules of \mnet and in~\autoref{sec:usecases}, we show how \mnet facilitates the generation of motion planning datasets incorporating new robots into existing scenes without much effort.
In~\autoref{subsec:hypothesis}, we show that it is possible to infer an incorrect conclusion when comparing motion planning algorithms due to limited data, emphasizing the importance of \mnet's problem generation.
In~\autoref{subsec:challenge} we show that the prefabricated datasets in \mnet are challenging even for fine-tuned planners, and no sampling-based planner rules over all.

%% file: src/related.tex
\input{imgs/table_related_work}

\section{Related Work}
\label{sec:related}

\begin{figure*} [ht!]
    \vspace{1em}
    \centering
    \includegraphics[width=0.95\textwidth]{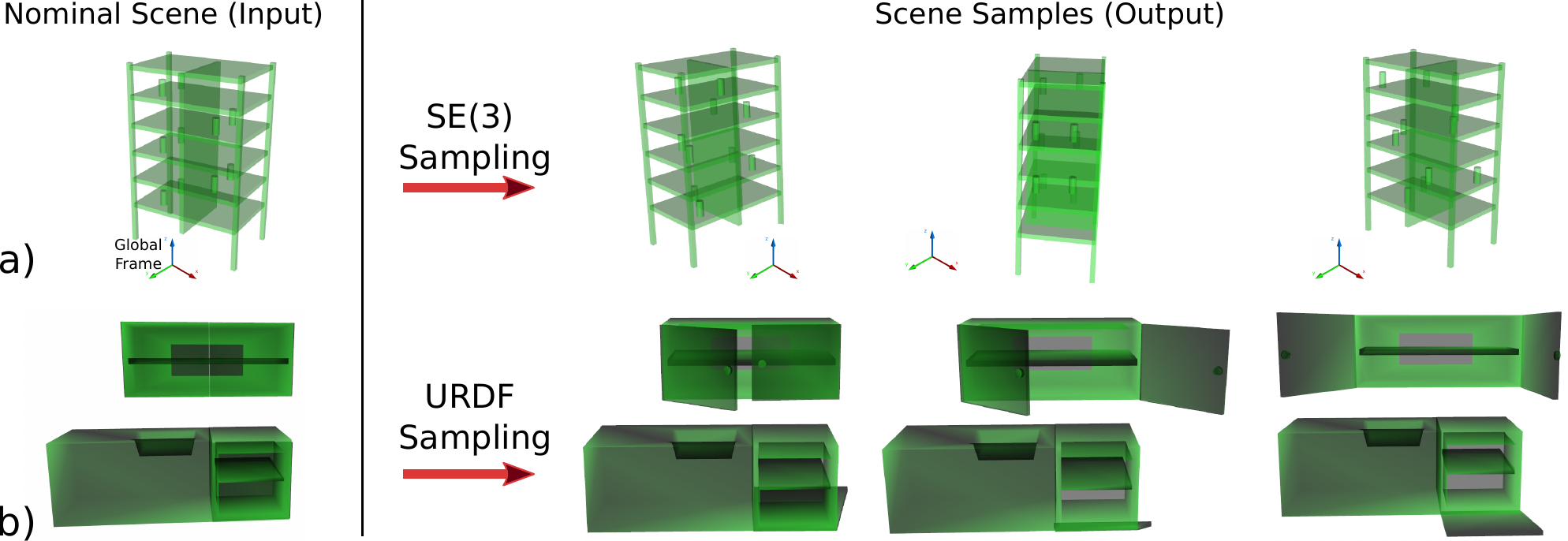}
    \caption{\ssampler: %
    The scene sampler module generates variations on a nominal input scene given sampling parameters.
    When performing \se sampling (shown in \textbf{a)}), variation in the pose of objects in the scene can be specified \emph{globally} (e.g., the bookshelf moving) and \emph{locally} (e.g., the cylinders moving on the shelves of the bookshelf).
    When performing URDF sampling (shown in \textbf{b)}), valid configurations of the kinematic structure specified by the URDF are sampled, which results in different configurations of the cabinets.
    }
    \vspace{-1em}
    \label{fig:scene_sampler}
\end{figure*}

Well-maintained datasets such as ImageNet~\cite{Deng2009imagenet} or Tencent ML-Images~\cite{Wu2019tencent} are fundamental for algorithmic breakthroughs in research fields like computer vision.
To achieve similar feats, the robotics community has developed several high-quality datasets.
We give a brief overview of the most popular ones (as of August 2021) with a focus on datasets for manipulation planning.
A more detailed overview can be found in~\cite{Calli2015benchmarking} (up until 2015). 

We compare datasets with each other based on six desirable properties.
First, we compare if a dataset is \emph{procedurally generated}, meaning if there exists an algorithmic generation of problems from a given scenario.
Second, we compare \emph{planner benchmarking} capabilities, meaning if there exists a tool to benchmark different motion planning algorithms on the dataset.
Third, if a dataset is procedurally generated, we check if there is an interface with \emph{tunable parameters}, i.e., if users can influence the generation process.
Fourth, we check if the dataset is \emph{high-dof}, i.e., if there exist robots with more than $6$-\dof.
Fifth, we check if the dataset contains \emph{sensed representations}, i.e., if there exist environments in the dataset which are built from sensor information.
Finally, we check for \emph{articulated robots}, i.e., if the datasets contain robots that are beyond rigid bodies in 2D or 3D.
Other properties could be examined, but we consider these properties necessary for a tool that focuses on manipulation.

As can be seen in~\autoref{tab:related_work}, we divide the datasets into three categories.
The first category is datasets for vehicle navigation.
Several high-quality datasets exists like common road~\cite{Althoff2017road}, bench mobile robot~\cite{Heiden2021} and the benchmark for autonomous robot navigation (BARN)~\cite{perille2020benchmarking}.
Similar datasets concentrate on indoor-navigation~\cite{RuizSarmiento2017}, 2D multi-agent path-finding~\cite{Stern2019mapf}, discrete point-robot path finding in 2D and 3D~\cite{Toma2021pathbench}, free-flying robots~\cite{moll2015benchmarking} or drones~\cite{Tani2020integrated}.
Our paper is complementary to vehicle navigation in that we concentrate on robot manipulation tasks.

The second category of datasets is focused on general robotics. These works aim at covering broad robotic categories like providing datasets and tools for remote teleoperation~\cite{Mandlekar2019} or object rearrangement~\cite{Liu2021ocrtoc}. While many papers are concentrating on learning-based approaches~\cite{James2020RLBench}, there is also a trend towards more reproducibility, for example by using containerization~\cite{Toma2021pathbench} to ease comparison over different operating systems or configurations. 

However, several tools in robotics have been developed specifically for manipulation tasks.
While the data generation is often similar, approaches differ by focusing either on learning-based algorithms or on planning-based algorithms.
In learning-based approaches like RobotNet~\cite{dasari2019robonet}, the focus is more on generating diverse camera streams.
In planning-based approaches like the Brown planning benchmark~\cite{Murray2020roadmap} the focus is more on creating mesh-based representations of the world useful to benchmark motion planners~\cite{Sucan2012}.
Other frameworks like ProbRobScene~\cite{Innes2021probrobscene} are independent of the algorithm used and focus instead on generating scenes automatically.
A particular dataset aimed at grasping is GraspNet, which concentrates on using the YCB dataset of objects~\cite{Fang2020graspnet} to generate large sets of grasping poses.
Similar datasets and benchmark utilities concentrate on specific aspects of manipulation. This involves tasks like bimanual manipulation~\cite{Chatzilygeroudis2020}, in-hand manipulation~\cite{Cruciani2020}, cloth manipulation~\cite{Garcia2020benchmarking}, aerial manipulation~\cite{Suarez2020benchmarks}, or solving Rubik's cube~\cite{Yang2020benchmarking}.

\mnet differs from all those approaches by (a) focusing on benchmarks specifically for motion planning algorithms, (b) having an incremental generation tool to create diverse sets of manipulation tasks, and (c) by concentrating on broad manipulation capabilities for diverse high-dimensional robotic arms.
This involves not only single gripper grasps but also bimanual manipulation (e.g., using the Baxter robot) and multi-finger manipulation (e.g., using the ShadowHand robot).

%% file: imgs/table_related_work.tex
\def\markYes{\textcolor{gg}{\checkmark}}
\def\markNo{\textcolor{gr}{$\times$}}
\def\markUndef{\textcolor{red}{E}}
\newcolumntype{Y}{>{\centering\arraybackslash}m{.001\linewidth}}
\newcolumntype{T}{>{\raggedright\arraybackslash}m{.05\linewidth}}

\newcommand{\addPlanner}[3]{%
#1 & #2 &
    \xintFor ##1 in {#3} \do {
        \xintifForFirst{}{&} 
        \ifnum##1=1
            \markYes
        \else
            \ifnum##1=-1
                \markUndef
            \else
                \markNo
            \fi
        \fi
    }
}

\newcommand\Tstrut{\rule{0pt}{2ex}}         
\newcommand\Bstrut{\rule[-1ex]{0pt}{0pt}}   
\newcommand{\addTopic}[1]{%
    \hline
    \multicolumn{8}{|c|}{#1}\Tstrut\Bstrut\\
    \hline
}
\begin{table}[t]
    \vspace{1em}
\caption{Relevant datasets in robotics as of August 2021.     \label{tab:related_work}}
    \footnotesize
    \center
    \begin{tabularx}{\linewidth}{| X | T | *{6}Y |}
         \hline
         Paper & Year & 
         \rotatebox{90}{Procedurally Generated} &  
         \rotatebox{90}{Planner Benchmarking} &  
         \rotatebox{90}{Tunable Parameters} &  
         \rotatebox{90}{High-Dof ($>6$)} &  
         \rotatebox{90}{Sensed Representation} &
         \rotatebox{90}{Articulated Robots} \\
\addTopic{Vehicle Navigation}
\addPlanner{CommonRoad \cite{Althoff2017road}}{2017}{0,1,0,0,0,0}\\
\addPlanner{Robot@Home \cite{RuizSarmiento2017}}{2017}{0,0,0,0,1,0}\\
\addPlanner{Multi-Agent Path-Find Benchmark\cite{Stern2019mapf}}{2019}{0,1,0,0,0,0}\\
\addPlanner{MAVBench \cite{Tani2020integrated}}{2020}{0,0,0,0,1,0}\\
\addPlanner{BARN \cite{perille2020benchmarking}}{2020}{1,1,1,0,0,0}\\
\addPlanner{Bench-MR \cite{Heiden2021}}{2021}{0,1,0,0,1,0}\\
\addPlanner{PathBench \cite{Toma2021pathbench}}{2021}{1,1,1,0,0,0}\\
\addTopic{General Robotics}
\addPlanner{OMPLBenchmarks \cite{moll2015benchmarking}}{2015}{0,1,0,0,0,0}\\
\addPlanner{Robobench \cite{Weisz2016robobench}}{2016}{0,1,0,1,1,1}\\
\addPlanner{Roboturk (Teleoperation database) \cite{Mandlekar2019}}{2019}{0,0,0,1,1,1}\\
\addPlanner{RLBench \cite{James2020RLBench}}{2020}{1,0,1,1,1,1}\\
\addPlanner{OCRTOC \cite{Liu2021ocrtoc}}{2021}{1,1,0,1,1,1}\\
\addTopic{Robot Manipulation}
\addPlanner{ACRV picking benchmark \cite{Leitner2017acrv}}{2017}{0,1,0,1,1,1}\\
\addPlanner{RoboNet \cite{dasari2019robonet}}{2019}{0,0,0,1,1,1}\\
\addPlanner{GraspNet \cite{Fang2020graspnet}}{2020}{0,0,0,0,0,0}\\
\addPlanner{Brown Planning Benchmarks \cite{Murray2020roadmap}}{2020}{1,1,0,1,0,1}\\
\addPlanner{Aerial Manipulation \cite{Suarez2020benchmarks}}{2020}{0,0,0,1,1,1}\\
\addPlanner{Bimanual Manipulation Benchmark \cite{Chatzilygeroudis2020}}{2020}{0,1,0,1,1,1}\\
\addPlanner{In-hand manipulation benchmark \cite{Cruciani2020}}{2020}{0,0,0,1,0,1}\\
\addPlanner{ProbRobScene \cite{Innes2021probrobscene}}{2021}{1,0,1,1,0,1}\\
\hline
\addPlanner{\mnet \textbf{(ours)}}{2021}{1,1,1,1,1,1}\\
    \hline
    \end{tabularx}
\end{table}

%% file: src/modules.tex
\section{Library Modules}
\label{sec:components}

\mnet is a flexible modular library composed of four basic modules, shown in~\autoref{fig:mnet_arch}.
The \ssampler shown in~\autoref{fig:scene_sampler}, creates variations of a given nominal scene.
The \ogen, shown in~\autoref{fig:octomap_generator}, converts a geometric scene to a point cloud and subsequently an octomap.
The \pgen generates motion planning problems given a scene, robot, and necessary configuration files. 
Finally, the \ssetup module enables the easy creation and usage of the generated datasets.

\subsection{Scene Sampler}

\label{subsec:scene_sampler}
Given variation parameters, the \ssampler module procedurally generates multiple scenes by randomly changing the nominal scene.
Currently, two complementary types of sampling are provided namely \emph{\se} and \emph{URDF} sampling as shown in~\autoref{fig:scene_sampler}.

\subsubsection{\se sampling}
For \se sampling, the nominal scene is a set of collision objects with \se poses relative to the global frame (shown in~\autoref{fig:scene_sampler}a). 
New scenes are generated by adding random noise to the \se poses of the collision objects~\cite{yershova2004deterministic}. 
The pose of the collision objects in the nominal scene serves as the mean of the sampling distribution and the variance (Gaussian) or bounds (Uniform) parameters are specified through a configuration file.
Finally, the random perturbations to the collision objects' poses can happen both  \emph{globally}, e.g., the shelf in~\autoref{fig:scene_sampler}a is moved with respect to the global frame, and \emph{locally}, e.g., the cylinders in~\autoref{fig:scene_sampler}a are perturbed with respect to the local frame of the shelf. Examples of samples drawn are shown on the right side of~\autoref{fig:scene_sampler}a.

\subsubsection{URDF sampling}
In this type of sampling, the nominal scene is specified as a URDF (Unified Robot Description Format) file~\cite{urdf}.
The URDF specifies the number of joints that describe the kinematic relations of objects in the scene.
By sampling valid configurations of this URDF (that is, collision-free with itself), we can generate different scenes.
This type of sampling emulates movements of objects subject to kinematic constraints such as cabinets opening and closing, shown in~\autoref{fig:scene_sampler}b. 

\subsection{Octomap Generator}
\label{subsec:octomap_generator}

\begin{figure}
    \centering
    \includegraphics[width=0.95\linewidth]{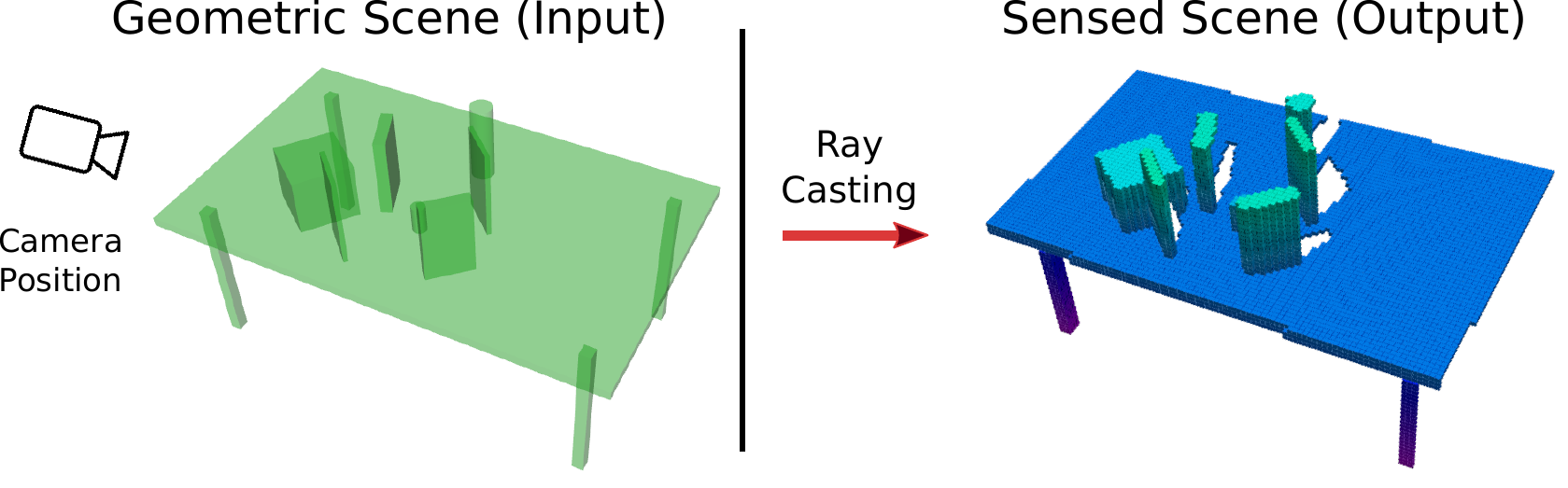}
    \caption{\ogen: The octomap generator module generates a sensed representation (point cloud and octomap) by emulating a depth camera from different positions.}
    \label{fig:octomap_generator}
\end{figure}

\ogen is an optional module that provides a way to convert geometric scene representations (i.e., geometric primitives and meshes) to point clouds and octomaps~\cite{Hornung2013}, as shown in~\autoref{fig:octomap_generator}.
The point cloud is generated by specifying in the frame(s) of the depth camera which is simulated with \texttt{gl\_depth\_sim}\footnote{\url{https://github.com/Jmeyer1292/gl_depth_sim}}.
This point cloud is later converted to an octomap. 
When a dataset is generated, all three representations (geometric, point cloud, octomap) can be simultaneously produced. 
Note that for motion planning, an octomap representation usually has a much higher collision checking time and is an over-approximation of the geometry,  leading to harder motion planning problems.
Nevertheless, we consider the sensed representation more ``realistic'' since it can be provided from any {RGB-D} camera, and is often used in practice.

\subsection{Problem Generator}
\label{subsec:problem_generator}
\begin{figure}
    \vspace{1em}
    \centering
    \includegraphics[width=0.95\linewidth]{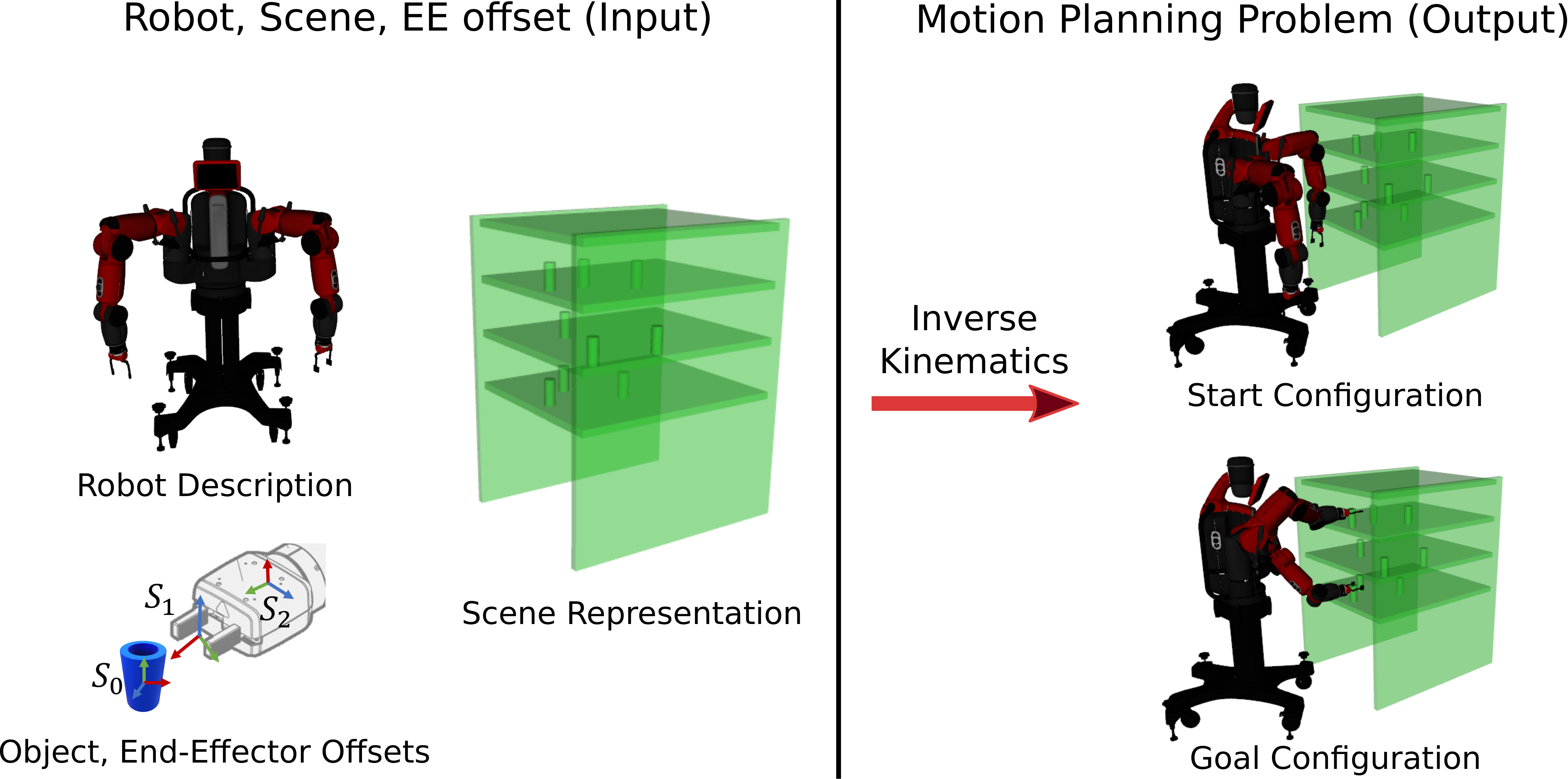}
    \caption{\pgen: %
    Given a robot description (URDF), geometric scene, and object-centric end-effector poses, the problem generator creates a motion planning problem.
    To have robot-agnostic problems, the start and goal of the problem can be specified as end-effector grasp poses, e.g., by specifying pose $S_1$ relative to the object's reference frame $S_0$. Additionally, a robot-specific transformation $S_2$ relative to the robot's end-effector frame is applied to the pose to account for different gripper idiosyncrasies. Full joint configurations can also be used as start/goal.  
Finally, a robot also has an optional base offset (not shown) that can be specified.
    }
    \label{fig:problem_generator}
\end{figure}

One critical idea in \mnet is the fact that motion planning problems can be easily generated for any robot-scene pair.
When done by hand, this process can be challenging and time-consuming since valid start and goal joint configurations in a motion planning problem depend on both the robot and the scene. 
The \pgen provides this functionality by defining a set of start and goal manipulation \emph{queries}. 
These \emph{queries} are specified as pose offsets expressed in the frame of collision objects in the nominal scene.
For example as seen in \autoref{fig:problem_generator}, the query expresses how the blue cup can be grasped. This is achieved by defining appropriate object-centric offsets that are robot agnostic.
This specification is conceptually similar to the affordance templates proposed by~\cite{hart2015affordance}.
Finally, objects in the scene can be attached to the end-effector(s) to emulate pick and place tasks. 

The \pgen creates full-motion planning requests (i.e., start/goal configurations in joint space) by performing collision-aware inverse kinematics.
These requests can be readily used together with a scene, to create a varied set of motion planning problems.
Note that there can be multiple queries defined for a scene, but the generated requests will consist of a single start-goal pair.
During generation, a planner can optionally be used to verify the feasibility of a problem.

As an additional feature, the \pgen supports the specification of manipulation queries for multiple end-effectors in the kinematic chain, e.g., multi-tip queries.
This is useful for applications in bimanual manipulation and when planning for dexterous hand robots such as the bookshelf with Baxter and the Shadowhand examples respectively (see~\autoref{fig:problems}).

\subsection{Setup}
\label{subsec:setup}

\begin{figure}
  \begin{center}
\begin{lstlisting}[language=C++]
 // Load the dataset given a meta-data file
 auto setup = std::make_shared<Setup>("conf.yaml");|\label{line:1}|
 auto robot = setup->getRobot();|\label{line:2}|
 auto planner = setup->createPlanner("planner");|\label{line:3}|
 Experiment experiment("exp", Profiler::Options());|\label{line:6}|
 
 for (int i = 1; i <= setup->getNumSamples(); i++)
 {
     // Load the ith scene in the dataset
     auto scene = std::make_shared<Scene>(robot);
     setup->loadGeometricScene(i, scene);|\label{line:4}|
 
     for (auto planner_name : {"PRM", "BiEST"})
     {
         // Load the start and goal configuration
         auto request = setup->createRequest();
         setup->loadRequest(i, request);|\label{line:5}|
 
         // Set planner.e.g., PRM, BiEST
         request->setConfig(planner_name);
         experiment.addQuery( //
           planner_name, scene, planner, request);|\label{line:7}|
     }
 }
 
 auto data = experiment.benchmark();|\label{line:8}|
 OMPLPlanDataSetOutputter output("results.log");
 output.dump(*data);

\end{lstlisting}
  \end{center}
  \caption{%
    A code snippet demonstrating how to load a dataset and benchmark different planners through the \ssetup module.  
}
\label{fig:setupcode}
\end{figure}

A convenient \ssetup class provides an easy-to-use interface to load created datasets and create \emph{planner}, \emph{scene} and \emph{robot}.
An example script with \ssetup is shown in \autoref{fig:setupcode}.
A dataset created by \mnet comes with a meta-data manifest (line~\autoref{line:1}, ``conf.yaml'').
This manifest contains all the relevant parameters that define the dataset and allow the user to access the sampled scenes and requests.
Once loaded, \ssetup can create instances of a \emph{robot} (line~\autoref{line:2}) and a \emph{planner} (line~\autoref{line:3}).
Our library takes advantage of the Robowflex~\cite{Robowflex} library to provide these constructs---Robowflex encapsulates the MoveIt~\cite{MoveIt} library for motion planning and provides capabilities for planning inside simple scripts.

The \ssetup class also provides a simple way to access each \emph{scene} (line~\autoref{line:4}) and corresponding \emph{request} (line~\autoref{line:5}) within a dataset.
After creating an \emph{experiment} (line~\autoref{line:6}), it is easy to add this specific problem (a \emph{scene} and \emph{request}, line~\autoref{line:7}) to the set of problems to benchmark.
After a benchmark is executed (line~\autoref{line:8}), the collected data can be output into a variety of formats, e.g., a SQL database compatible with PlannerArena~\cite{moll2015benchmarking}. 

%% file: src/usecases.tex
\section{Example Usecases}
\label{sec:usecases}
\begin{figure*}[h]
         \vspace{1em}
         \centering
         \includegraphics[width=\linewidth]{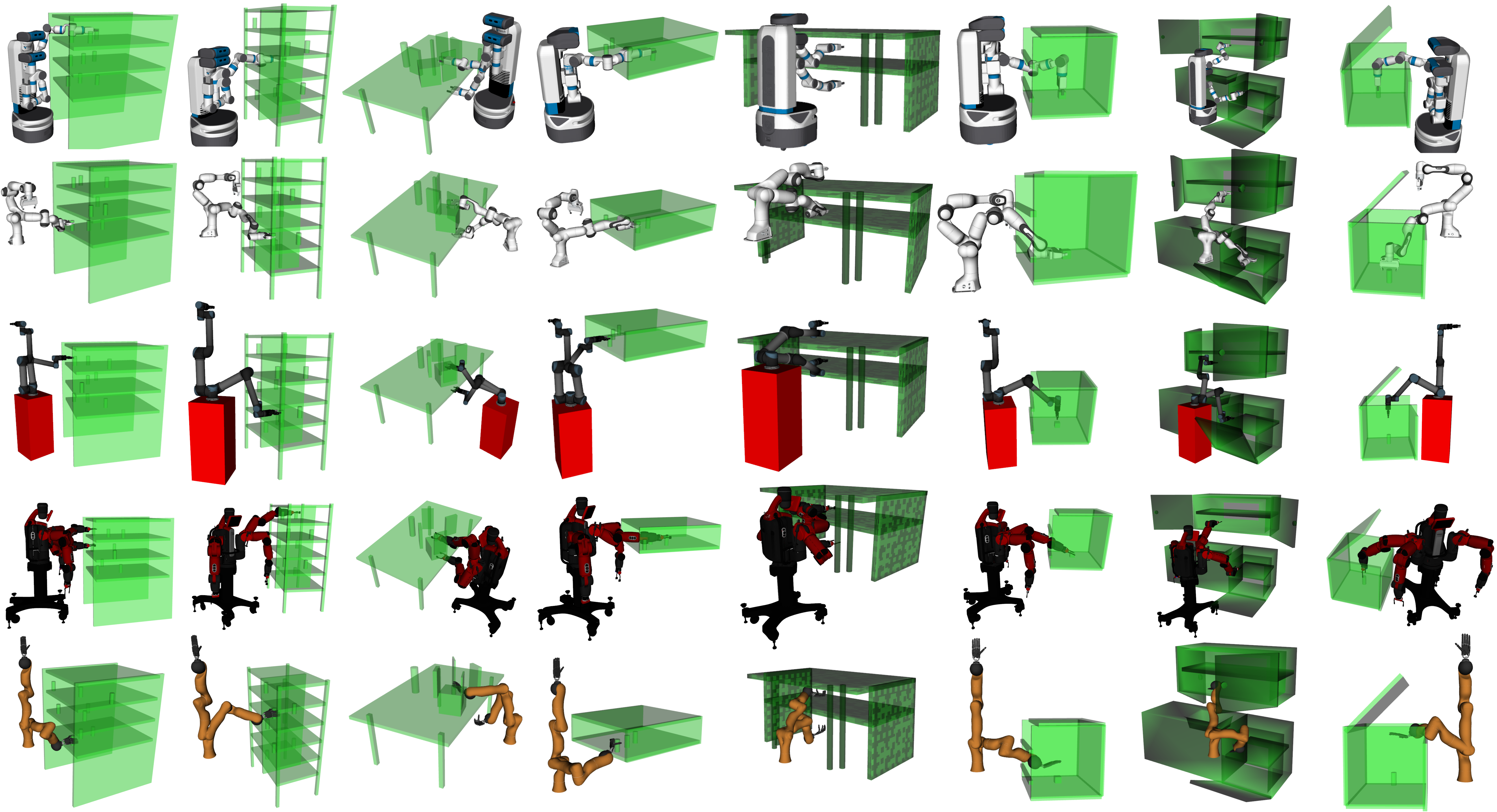}
         \caption{%
         Representative problems from the $40$ different prefabricated datasets provided with \mnet.
         There are $8$ nominal scenes and $5$ robots, which create the $40$ datasets, each consisting of 100 different problems.
         } 
         \label{fig:problems}
         \vspace{-1em}
\end{figure*}

The user interacts with \mnet in two ways: by creating C++ scripts that call library modules or by specifying values in configuration files to define new problems.
The first case of using C++ scripts was shown in~\autoref{subsec:setup}.
There (shown in \autoref{fig:setupcode}) the user loads an existing dataset in \mnet to benchmark different motion planners--- benchmarking results can be plotted and analyzed through PlannerArena~\cite{moll2015benchmarking}.

The second case considered is a user who desires to create a new dataset with a robot or scene not currently in \mnet.
The user simply needs to provide a robot description and scene description file along with the required offsets (\autoref{subsec:problem_generator}).
Given these files, \mnet through a script will procedurally generate varied motion planning problems, without the burden of manually creating valid start/goal configurations and scene samples for different problems.
For example, we used this script for the $40$ different prefabricated datasets, shown in~\autoref{fig:problems}.
We created $8$ nominal scenes and specified the end-effector and base offsets of the following $5$ robots: a Fetch ($7$-$8$-\dof) a Panda ($7$-\dof), a UR5 ($6$-\dof), a Baxter ($7$-$14$-\dof) and a ShadowHand mounted on a KUKA arm ($31$-\dof).

To verify that each generated problem is feasible, we used a highly-tuned sampling-based planner with a large timeout (60 seconds) and discarded problems that could not be solved in time.
For each dataset, an arbitrary number of motion planning problems can be generated but for our purposes, we created $100$ motion planning problems for each dataset.

Finally, \mnet has already been used to create a diverse set of datasets suitable for learning-based methods~\cite{Chamzas2020sampling, pairet2021}, for hyper-parameter tuning methods~\cite{moll2021hyperplan}, for planning under uncertainty~\cite{quintero2021robust}, for planning in partially observable environments~\cite{quintero2021blind} and for planning on different abstraction levels~\cite{Orthey2020, Orthey2021TRO}.

%% file: src/experiments.tex
\section{Evaluations}
\label{sec:experiments}
In this section, we present two evaluations to showcase the efficacy of \mnet.
In \autoref{subsec:hypothesis} we demonstrate how using few motion planning problems can potentially lead to wrong conclusions, for example when comparing two motion planners.
In \autoref{subsec:challenge} we demonstrate that many of the prefabricated datasets are challenging and no specific planner outperforms the other ones.

\begin{figure*}
    \vspace{1em}
    \centering
    \includegraphics[width=0.95\linewidth]{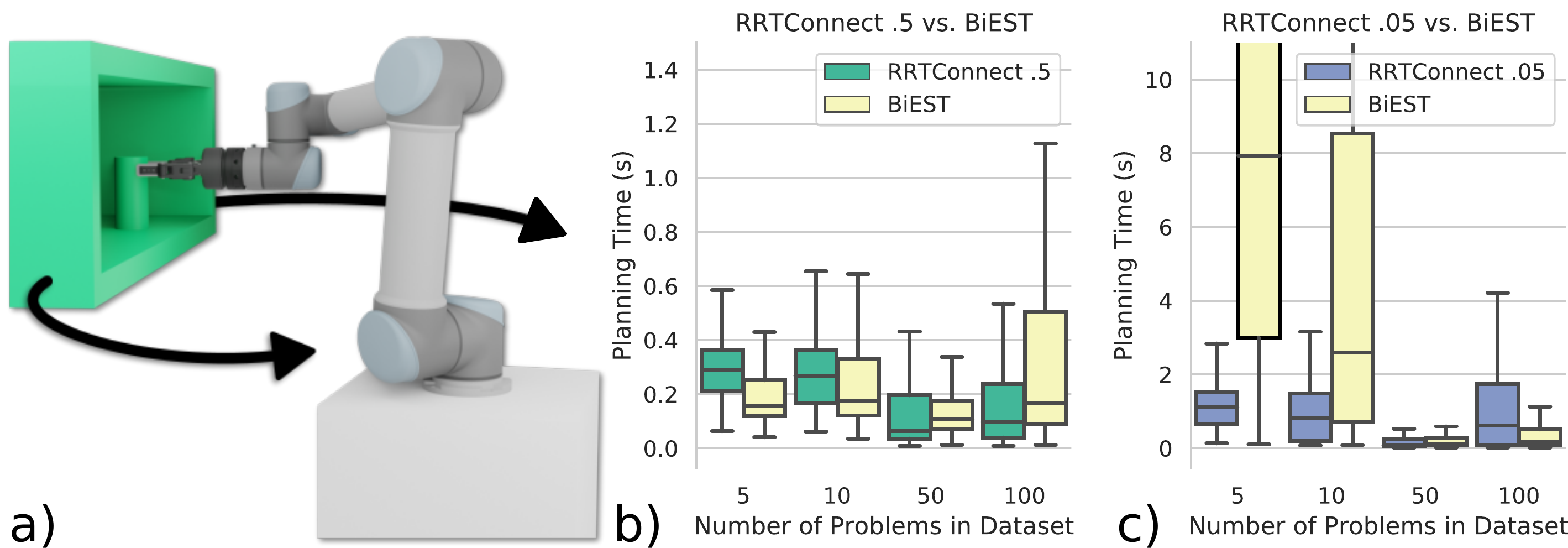}
    \caption{
    \textbf{a)} One of the 100 sampled problems in a simple shelf-picking dataset for a UR5 robot.
    Both the relative angular position of the shelf as well as the position of the cylinder (not shown for visual clarity) vary between motion planning problems.
    \textbf{b), c)} Two different adversarial orderings of the same 100 motion planning problems.
    The \yaxis shows the planning time while the \xaxis is the number of motion planning problems considered.
    A timeout of 60 seconds was used for all the problems and each motion planning problem was solved 20 times.
    It is clear that when using only a small number of motion planning problems, the wrong conclusions can be drawn, as the best performing planner can be different when considering all 100 problems.
    }
    \vspace{-1em}
   \label{fig:hyp_time}
\end{figure*}

\begin{figure}
  \centering
  \includegraphics[width=0.95\linewidth]{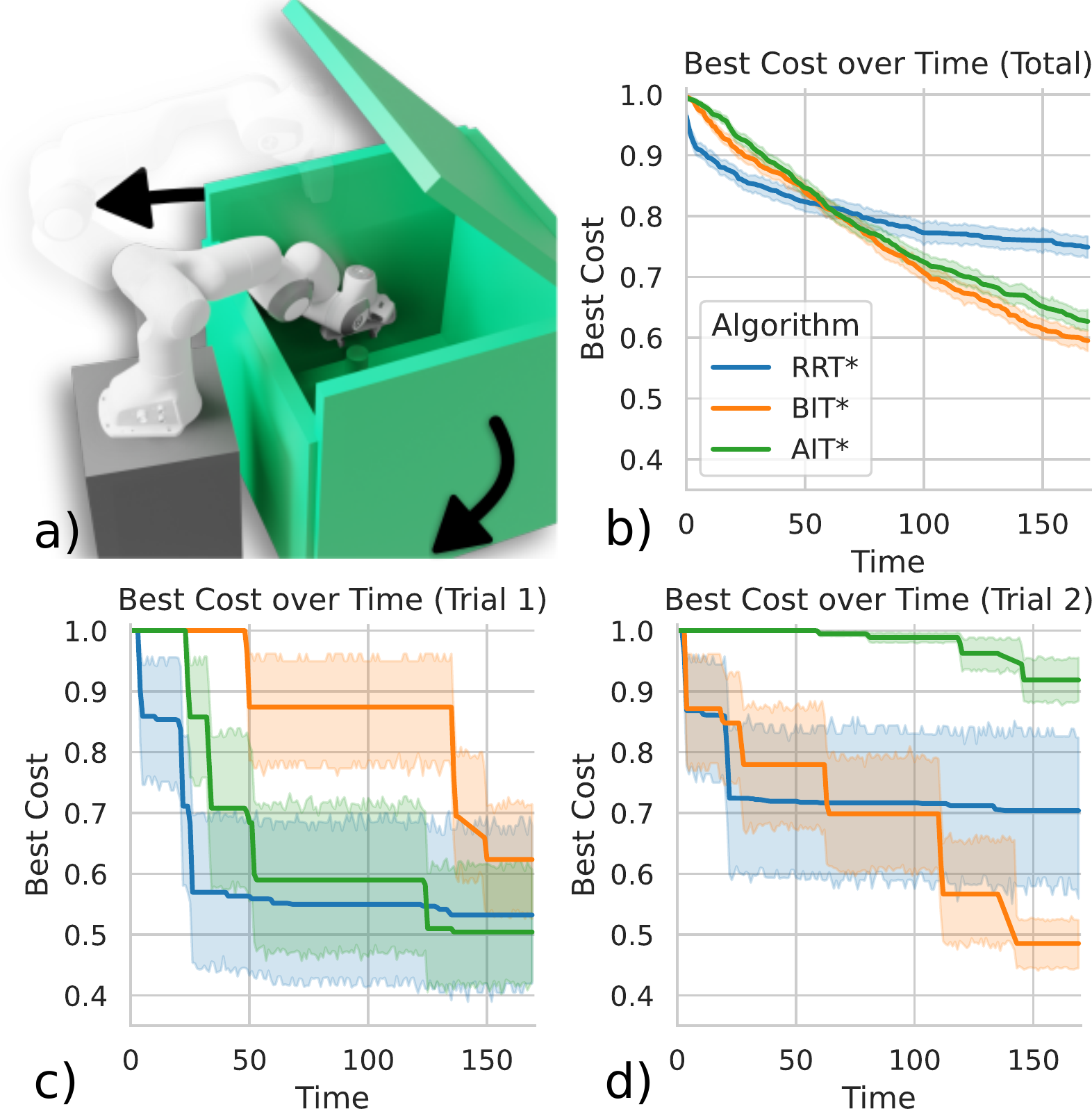}
  \caption{%
  \textbf{a)} One of the 100 sampled in a box-picking dataset with the Panda robot.
  In this task, both the relative angular position of the box and the position of the object in the box (not shown for visual clarity) vary.
  \textbf{b), c), d)} Show the normalized cost for 3 different optimizing planners in the box-picking task.
  The median is shown of 5 independent runs for each planner with a $99\%$ confidence interval.
  Trial 1 and 2 shown in \textbf{c), d)} show the convergence plots of \rrts \bit and \ait on a single, specific motion planning problem while \textbf{b)} 
  shows results considering all 100 problems.
  Results in aggregate differ between Trial 1 and Trial 2 demonstrating that using a few motion planning problems could potentially lead to incorrect conclusions.
  }  
  \label{fig:hyp_length}
\end{figure}

\begin{figure*}
         \vspace{1em}
         \centering
         \includegraphics[width=0.95\linewidth]{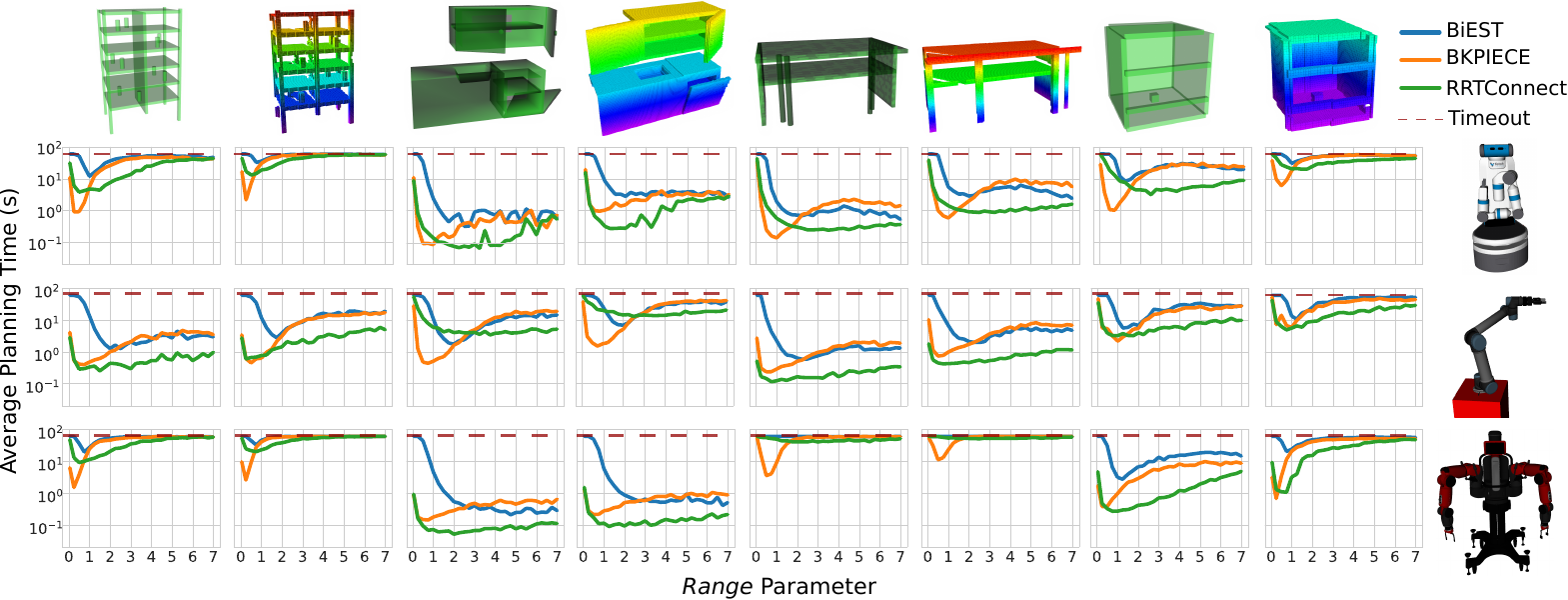}
         \caption{%
             Timing results for three planners (\bkpiece,  \rrtc, and \biest) in 4 environments from~\autoref{fig:problems} (both geometric and sensed) on 3 different robots (the Fetch, UR5, Baxter) for a total of 12 datasets.
             We plan for both arms of the Baxter in the table environment while for only in the rest.
             In each plot on this matrix, the value of each of the planner's \emph{range} parameters (which controls \cspace expansion) is varied between 0 to 7, in increments of 0.25.
         The average planning time over the 100 problems in the dataset for each range parameter is shown on a log scale, plotted as a line.
         Planning timeout was 60 seconds, visualized as a dotted red line---lines touching the red line indicate planning timeout.
         Note that for many problems, specific tunings of the planners are required to solve the problem, while no planner consistently outperforms all others,  indicating their difficulty and diversity.
         } 
         \vspace{-1em}
         \label{fig:challenge}
\end{figure*}

\subsection{Wrong Hypothesis}
\label{subsec:hypothesis}

Undoubtedly, in any research field, it is necessary to compare the performance of different methods.
In motion planning research, it is often the case that a practitioner has a specific robot and target application in mind, which begets the need to manually construct an appropriate benchmark.
Creating a benchmark from scratch without the appropriate tools is both time-consuming and challenging since a large number of problems might be required to achieve statistical significance.  
We note here that the designed experiments highlight the importance of using a  large number of problems when comparing different planners and should not be interpreted as an indication of which planner is best.
Unless indicated, planners are using default parameters from OMPL~\cite{Sucan2012}.

Consider the following hypothetical scenario: a practitioner wants to compare different planners on a picking task.
Specifically, the problem of interest involves a UR5 robot tasked with picking a cylinder from a shelf, shown in~\autoref{fig:hyp_time}a.
Say the practitioner either samples or chooses specific instances of this scene: in~\autoref{fig:hyp_time}b and \autoref{fig:hyp_time}c, we present the worst case scenarios for this practitioner (for 5, 10, 50, and 100 problem instances) in terms of drawing conclusions on their planner's performance.

We generated 100 feasible motion planning problems as described in~\autoref{sec:usecases} and benchmarked planning time for \biest~\cite{Hsu1998} and \rrtc~\cite{LaValle2000} (we use \rrtc with two different ``range'' values, 0.05 and 0.5, which controls the \cspace expansion step).
For each specific problem (an instance of the scene), the problem was solved 20 times (with a 60 seconds timeout), for a total of 2000 data points given 100 scenes.
In figures \autoref{fig:hyp_time}b and \autoref{fig:hyp_time}c you can see two different adversarial orderings of the data.
That is, for both of these plots, we sorted the same motion planning problems in the dataset such that problems early in the dataset have the largest difference in average planning time between the two compared planners.
In \autoref{fig:hyp_time}b \biest and \rrtc with range $0.5$ are compared.
The \xaxis denotes how many problems from the sorted problems are considered.
Here, \biest is better when considering only 5 or 10 problems, while when considering the entire dataset (100 problems) it is clear that \rrtc is more performant.
In \autoref{fig:hyp_time}c, the same effect is demonstrated between \biest and \rrtc with a range parameter of $0.05$, with \biest faster only after aggregating the results from all 100 problems. 
This empirically shows the danger in considering only a few problem instances for evaluation.
\mnet provides the tools necessary to easily create varied datasets to help avoid this problem.

Beyond planning time, this phenomenon could occur when comparing other planner metrics, e.g., comparing the best cost over time for asymptotically-optimal sampling-based planners, as shown in~\autoref{fig:hyp_length}.
Here the experiment entails a Panda robot grasping a cylinder from the box with similar variation as in~\autoref{fig:hyp_time}.
In this example cost is defined as joint path length, but different costs suchs as clearance or cartesian length can be specified through the MoveIt \cite{MoveIt} interface. 
We show the median of the best normalized cost found for \rrts~\cite{Karaman2011}, \bit~\cite{Gammell2015}, and \ait~\cite{strub2020adaptively}.
Each planner is run 5 times per problem, with a given 180 seconds planning time.
\autoref{fig:hyp_length}c and \autoref{fig:hyp_length}d indicate two different conclusions about which planner performs best when considering a single problem instance.
As above, incorrect conclusions would be drawn about planner performance in this domain if only based on a specific problem instance---\autoref{fig:hyp_length}b shows the aggregated results over all 100 problems, which provides a stronger conclusion.
Note that in general all datasets are prone to bias, but procedurally generating more instances ameliorates this bias.

\subsection{Benchmarking Results of Datasets}
\label{subsec:challenge}

In this section, we analyze the results of benchmarking 12 out of the 40 datasets on both the geometric and sensed (octomap) representations to demonstrate the difficulty of the provided datasets.
We benchmarked three bidirectional tree-based planners, namely \biest~\cite{Hsu1999}, \rrtc~\cite{LaValle2000}, and \bkpiece~\cite{Sucan2008} for different values of their range parameter as shown on the \xaxis of each subplot in \autoref{fig:challenge}.
We choose these planners, as among sampling-based planners they are typically highly performant in such tasks. 
Additionally, the \emph{range} parameter (used by each planner to control the rate of expansion in \cspace) is empirically known to have a significant influence on planning performance.  

Results are shown in \autoref{fig:challenge}.
We first note that these problems demonstrate a broad range of planning performance---each of these planners varies in performance according to environment and robot and there is no clear winner across the full spectrum of problems.
In several cases, even the most performant planner has more than 1 second of average planning time indicating the difficulty of the datasets. Moreover, note that planner performance is comparable between the geometric and the sensed problems, with a small performance hit in the sensed representation.
Finally, we verify that these planners are sensitive to the \emph{range} parameter, as there are clear performance peaks for the planners at specific range values for different problems.

%% file: src/discussion.tex
\section{Discussion}
In this paper, we have presented \mnet, a new open-source tool to procedurally generate and benchmark motion planning datasets.
\mnet supports a robot-agnostic specification of environments, sampling new planning problems from a specified distribution, and can generate ``sensed'' representations for realistic, challenging problems.
Through our experiments, we show the importance of procedurally generating datasets, as using only a few hand-designed problems could potentially lead to incorrect conclusions.

In the future, we would like to continue extending the repository of generated datasets with the help of the community, with more robots and environments as well as supporting sequential motion planning problems, such as in task and motion planning. 
We would also like to add features that help users profile their dataset with a set of metrics or features, e.g., space expansiveness, to help understand what are the challenging aspects of the proposed problem.
Some of the limitations of this work are that \se and \emph{URDF} sampling are only approximations of the variability of the real world, no camera data can be given to the planner for visual planning, and only geometric constraints are considered.
We hope to continuously improve this tool and that it will help the community advance the field of motion planning by supporting researchers to design and share benchmarking datasets.